\documentclass[10pt,letterpaper,twocolumn]{article}
\usepackage[latin9]{inputenc}
\usepackage{array}
\usepackage{booktabs}
\usepackage{multirow}
\usepackage{amsmath}
\usepackage{amssymb}
\usepackage{graphicx}
\usepackage[unicode=true,
 bookmarks=false,
 breaklinks=true,pdfborder={0 0 1},backref=section,colorlinks=false]
 {hyperref}
\hypersetup{
 pagebackref=true,letterpaper=true,colorlinks}

\makeatletter

\pdfpageheight\paperheight
\pdfpagewidth\paperwidth

\providecommand{\tabularnewline}{\\}

\usepackage{cvpr}
\usepackage{times}
\usepackage{epsfig}
\usepackage{graphicx}
\usepackage{caption}

\cvprfinalcopy 

\def\cvprPaperID{7883} 

\makeatother

\begin{document}
\title{Learning Asynchronous and Sparse Human-Object Interaction in Videos}
\author{Romero Morais$^{*}$, Vuong Le, Svetha Venkatesh, Truyen Tran\\
 Applied Artificial Intelligence Institute, Deakin University, Australia\\
 \texttt{\small{}\{ralmeidabaratad,vuong.le,svetha.venkatesh,truyen.tran\}@deakin.edu.au}{\small{}
}}

\twocolumn[{%
\renewcommand\twocolumn[1][]{#1}%
\vspace{-4em}
\maketitle
\vspace{-3em}
\begin{center}
    \centering
    \includegraphics[width=1\linewidth]{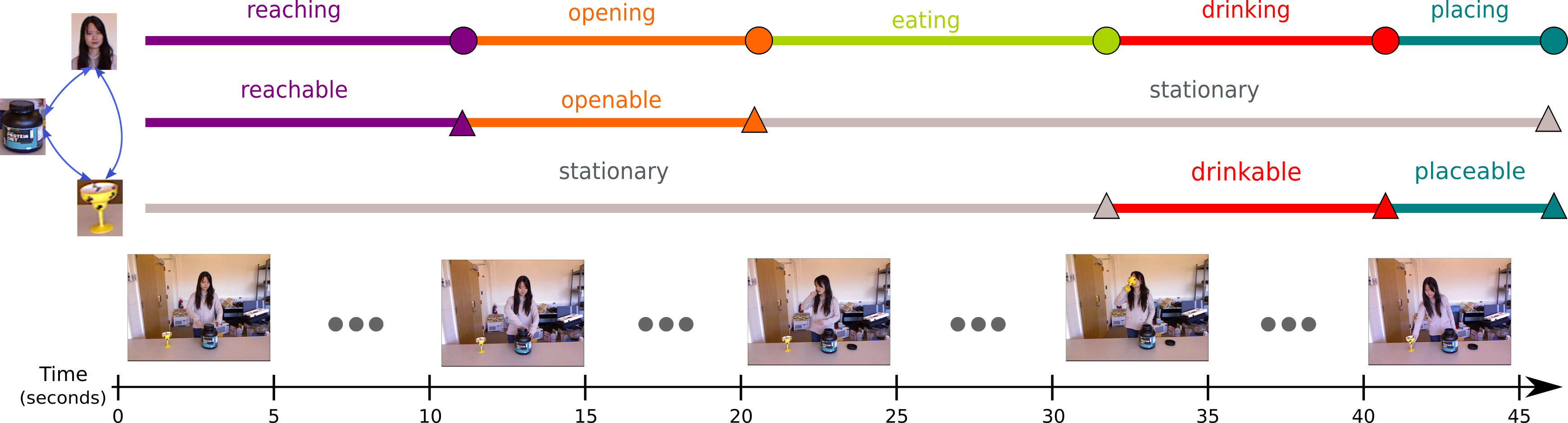}
    \captionof{figure}{An example of human-object interaction activity, where a person takes
some medicine and interacts with two objects. Human (circles) and
object (triangles) entities have independent lives throughout the
video (upper three rows). Although videos are captured in regular
timing (lower rows), the dynamics of human activities and object affordance
evolve sparsely and asynchronous with each other (colored segments).
They also affect each other (blue curved arrows). These characteristics
of human-object interactions are the main modeling goals of this work.\label{fig:paper-teaser}}
\end{center}%
}]

\maketitle

\global\long\def\model{\textrm{{Asynchronous-Sparse\ Interaction\ Graph\ Networks}}}%

\global\long\def\modelshort{\textrm{{ASSIGN}}}%

\global\long\def\intrasymbol{\textrm{intra}}%

\global\long\def\intersymbol{\textrm{inter}}%

\global\long\def\palemustardbox{\fcolorbox[cmyk]{0.0,0.0,0.0,0.0}{0.15,0.10,0.3,0.0}{\rule[3pt]{0pt}{1pt}\rule[3pt]{1pt}{0pt}}}%

\global\long\def\darkbluebox{\fcolorbox[cmyk]{0.0,0.0,0.0,0.0}{1.0,0.7,0.3,0.0}{\rule[3pt]{0pt}{1pt}\rule[3pt]{1pt}{0pt}}}%

\global\long\def\lightbrownbox{\fcolorbox[cmyk]{0.0,0.0,0.0,0.0}{0.0,0.063,0.255,0.341}{\rule[3pt]{0pt}{1pt}\rule[3pt]{1pt}{0pt}}}%

\global\long\def\orangebox{\fcolorbox[cmyk]{0.0,0.0,0.0,0.0}{0.0,0.5,1.0,0.0}{\rule[3pt]{0pt}{1pt}\rule[3pt]{1pt}{0pt}}}%

\global\long\def\lightbluebox{\fcolorbox[cmyk]{0.0,0.0,0.0,0.0}{0.392,0.0,0.0,0.0}{\rule[3pt]{0pt}{1pt}\rule[3pt]{1pt}{0pt}}}%

\global\long\def\lightpinkbox{\fcolorbox[cmyk]{0.0,0.0,0.0,0.0}{0.03,0.3,0.0,0.0}{\rule[3pt]{0pt}{1pt}\rule[3pt]{1pt}{0pt}}}%

\global\long\def\graybox{\fcolorbox[cmyk]{0.0,0.0,0.0,0.0}{0.0,0.0,0.0,0.07}{\rule[3pt]{0pt}{1pt}\rule[3pt]{1pt}{0pt}}}%

\global\long\def\redbox{\fcolorbox[cmyk]{0.0,0.0,0.0,0.0}{0.2,1.0,1.0,0.0}{\rule[3pt]{0pt}{1pt}\rule[3pt]{1pt}{0pt}}}%

\global\long\def\purplebox{\fcolorbox[cmyk]{0.0,0.0,0.0,0.0}{0.4,0.5,0.0,0.0}{\rule[3pt]{0pt}{1pt}\rule[3pt]{1pt}{0pt}}}%

\global\long\def\palebluebox{\fcolorbox[cmyk]{0.0,0.0,0.0,0.0}{0.3,0.15,0.1,0.0}{\rule[3pt]{0pt}{1pt}\rule[3pt]{1pt}{0pt}}}%

\global\long\def\paleorangebox{\fcolorbox[cmyk]{0.0,0.0,0.0,0.0}{0.0,0.2,0.5,0.0}{\rule[3pt]{0pt}{1pt}\rule[3pt]{1pt}{0pt}}}%

\global\long\def\darkpinkbox{\fcolorbox[cmyk]{0.0,0.0,0.0,0.0}{0.1,0.7,0.0,0.0}{\rule[3pt]{0pt}{1pt}\rule[3pt]{1pt}{0pt}}}%

\global\long\def\mediumbluebox{\fcolorbox[cmyk]{0.0,0.0,0.0,0.0}{0.314,0.047,0.0,0.118}{\rule[3pt]{0pt}{1pt}\rule[3pt]{1pt}{0pt}}}%

\global\long\def\palegreenbox{\fcolorbox[cmyk]{0.0,0.0,0.0,0.0}{0.043,0.0,0.321,0.11}{\rule[3pt]{0pt}{1pt}\rule[3pt]{1pt}{0pt}}}%

\global\long\def\darkgraybox{\fcolorbox[cmyk]{0.0,0.0,0.0,0.0}{0.0,0.0,0.0,0.7}{\rule[3pt]{0pt}{1pt}\rule[3pt]{1pt}{0pt}}}%

\global\long\def\darkbrownbox{\fcolorbox[cmyk]{0.0,0.0,0.0,0.0}{0.5,0.7,1.0,0.0}{\rule[3pt]{0pt}{1pt}\rule[3pt]{1pt}{0pt}}}%

\global\long\def\lightredbox{\fcolorbox[cmyk]{0.0,0.0,0.0,0.0}{0.0,0.5,0.5,0.0}{\rule[3pt]{0pt}{1pt}\rule[3pt]{1pt}{0pt}}}%

\global\long\def\lightgreenbox{\fcolorbox[cmyk]{0.0,0.0,0.0,0.0}{0.4,0.0,0.4,0.0}{\rule[3pt]{0pt}{1pt}\rule[3pt]{1pt}{0pt}}}%

\global\long\def\greenbox{\fcolorbox[cmyk]{0.0,0.0,0.0,0.0}{0.7,0.1,1.0,0.0}{\rule[3pt]{0pt}{1pt}\rule[3pt]{1pt}{0pt}}}%

\global\long\def\lightpurplebox{\fcolorbox[cmyk]{0.0,0.0,0.0,0.0}{0.25,0.35,0.0,0.0}{\rule[3pt]{0pt}{1pt}\rule[3pt]{1pt}{0pt}}}%

\global\long\def\palegreenbox{\fcolorbox[cmyk]{0.0,0.0,0.0,0.0}{0.25,0.15,0.5,0.0}{\rule[3pt]{0pt}{1pt}\rule[3pt]{1pt}{0pt}}}%

\global\long\def\pinksalmonbox{\fcolorbox[cmyk]{0.0,0.0,0.0,0.0}{0.03,0.3,0.2,0.0}{\rule[3pt]{0pt}{1pt}\rule[3pt]{1pt}{0pt}}}%

\global\long\def\turquoisebox{\fcolorbox[cmyk]{0.0,0.0,0.0,0.0}{0.5,0.0,0.15,0.0}{\rule[3pt]{0pt}{1pt}\rule[3pt]{1pt}{0pt}}}%

\begin{abstract}
Human activities can be learned from video. With effective modeling
it is possible to discover not only the action labels but also the
temporal structures of the activities such as the progression of the
sub-activities. Automatically recognizing such structure from raw
video signal is a new capability that promises authentic modeling
and successful recognition of human-object interactions. Toward this
goal, we introduce $\model$ ($\modelshort$), a recurrent graph network
that is able to automatically detect the structure of interaction
events associated with entities in a video scene. $\modelshort$ pioneers
learning of autonomous behavior of video entities including their
dynamic structure and their interaction with the coexisting neighbors.
Entities' lives in our model are asynchronous to those of others therefore
more flexible in adaptation to complex scenarios. Their interactions
are sparse in time hence more faithful to the true underlying nature
and more robust in inference and learning. $\modelshort$ is tested
on human-object interaction recognition and shows superior performance
in segmenting and labeling of human sub-activities and object affordances
from raw videos. The native ability for discovering temporal structures
of the model also eliminates the dependence on external segmentation
that was previously mandatory for this task.

\end{abstract}

\section{Introduction}

Human activities are strongly connected to the surrounding environment
and the objects in it. The interactions between human and object entities
observed in videos are a fundamental clue toward a deep understanding
of human behavior and the surrounding world \cite{gupta2009observing}.
This capability is reflected in the \emph{human-object interaction
(HOI) recognition} task, in which human sub-activities (such as ``drinking'')
and object affordances (such as ``drinkable'') are segmented and
recognized from a video by analyzing the interactive relations between
them. See Fig. \ref{fig:paper-teaser} for an example. These relations
naturally form a spatio-temporal graph in which entities (humans or
objects) and their dynamic interactions evolve throughout the activity.
Although entities can be detected and tracked from video, it is challenging
to build a graph model that can automatically discover the temporal
structure of activities and natively reflect the complex and intricate
nature of these interactions.

Currently available approaches applied conditional random field \cite{jiang2014modeling,koppula2013learning,koppula2016anticipating}
and graph neural networks \cite{ghosh2020stacked,qi2018learning}
to model the spatio-temporal entity interaction graph. These models
assumed knowledge of the temporal structure of the video, and limited
the task to assigning activity and affordance labels to the segments.

Rather than this cascading approach, we exploit the fact that the
structure and the content of the events are tightly coupled and may
support each other toward the optimal solution in a joint discovery
scheme. Such schemes further make possible to break the common assumption
that entities in a video are always active and their interactions
happen continuously. In reality, unlike regularly captured video frames,
interactions between entities happen sparsely in time. This suggests
that the temporal relations in the interaction graph can be pruned
for a more concise and efficient communication. Furthermore, while
bounding boxes are tracked from video frames concurrently, entities'
lives are asynchronous with each other. Authentic modeling of this
asynchronicity provides a great flexibility to allow entities to act
independently and only change their states when necessary.

We introduce Asynchronous-Sparse Interaction Graph Networks ($\modelshort$),
a joint structure-content discovery framework for sparse and asynchronous
human-object interactions. $\modelshort$ stands on the principle
that each entity has an independent life in a video, where it behaves
and interacts with its coexisting neighbors in its own pace and timing.
This flexible temporal structure and the content labels of events
are discovered jointly using two-layer dynamic graph networks that
can do inference and be trained end-to-end without dependence on external
segmentation labels or preprocessing.

The segmentation and labeling capabilities of $\modelshort$ are demonstrated
on two major human-object interaction datasets, with superior quantitative
performance and cleaner, more realistic qualitative results than currently
available methods. Furthermore $\modelshort$ shows the unique advantage
in modeling the interaction across multiple human entities on the
appropriate dataset.

In summary, this paper makes three major contributions:
\begin{itemize}
\item Constructs the first end-to-end graph model that jointly learn temporal
structure and content label of human-object interaction activities;
\item Effectively models the sparse and asynchronous entities lives in the
social context of activity video; and
\item Permits efficient relational inference that can skip unnecessary operations
resulting it in being robust to a wide variety of event structures.
\end{itemize}

\section{Related work}

\begin{figure*}[!tp]
\centering{}\includegraphics[width=1\textwidth]{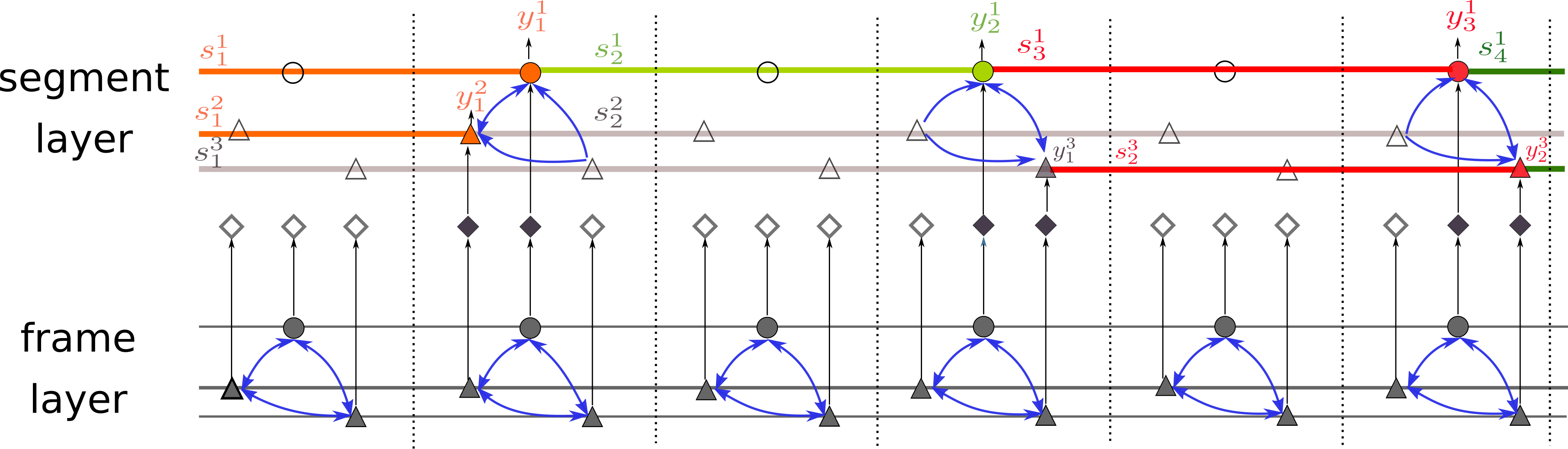}\caption{$\protect\model$ ($\protect\modelshort$) architecture contains two
layers of spatio-temporal graph networks. At each layer, graph nodes
represent human (circle) or object (triangle) entities. Spatial edges
are modeled by message passing (blue curved arrows), and temporal
edges are modeled through recurrent networks (horizontal lines). The
frame level of $\protect\modelshort$ updates at every time step for
every entity, and decides at each step (upward arrows) whether the
corresponding segment-level entity \emph{change} (solid diamond) or
\emph{skip} (hollow diamond)$\:$-$\:$details in Sec.\ref{subsec:Segmenting-the-entity}.
The sparse \emph{change} signals lead to asynchronous and sparse updates
(solid shapes) and interactions (blue curved arrows) at the segment
level of $\protect\modelshort$$\:$-$\:$details in Sec.\ref{subsec:Labeling-the-learned}.
The labels of segments are generated by the second layer at the update
operators.\label{fig:model-architecture}}
\end{figure*}

\subsection{Human-object interaction in videos}

Traditional approaches to HOI modeling in videos surround on variations
of Markov Random Fields (MRF). Koppula $\etal$ \cite{koppula2013learning}
used an MRF to model entities in videos with fully connected spatial
and temporal edges. It also starts a trend to use sub-activity segments
as temporal time units. This work is extended into the ATCRF model
\cite{koppula2016anticipating} that anticipates future sub-activity/affordances
and gathers features from frame-level nodes. ATCRF is further advanced
into GP-LCRF \cite{jiang2014modeling} to reduce the dimensionality
of the frame-level human representation. Another extension of ATCRF
is the Recursive CRF \cite{sener2015rCRF}, in which the CRF is placed
under a Bayesian filtering with an efficient belief computation. With
the recent advancement of spatio-temporal relation modules, MRF-like
models advanced into more efficient implementations with recurrent
neural networks (RNNs) and graph neural networks (GNNs). Jain $\etal$
\cite{jain2016structural-rnn} proposed to factorize the dynamic relationships
in HOI and model the factors with a mixture of RNNs. Qi $\etal$ \cite{qi2018learning}
proposed Graph Parsing Neural Network (GPNN), which allows spatial
graph topology to be inferred adaptively. Ghosh $\etal$ \cite{ghosh2020stacked}
extended GNNs with a stacked hourglass network \cite{newell2016stacked}
for label prediction. The MRF and GNN families of models are reliable
in predicting HOI labels but they cannot perform temporal segmentation
themselves and need to resort to an external segmentation (such as
dynamic programming) either before or during inference. $\modelshort$
on the other hand, learns the segmentation in tandem with the segment
labeling directly from frame-level features.

Such joint capability is also the goal of a couple of efforts to constrain
HOIs with activity grammars borrowed from natural language processing
namely Stochastic grammar \cite{qi2017predicting} and Earley tree
parser \cite{qi2018generalized}. These constraints improve the ability
to learn relationships between the entities, but at the same time
limits their flexibility in the process.

Shared between all previous works is the assumption that entities
are synchronous and constantly update their states. This oversimplification
is unnatural and being a source of practical issues such as over-segmentation.
In this work, we directly challenge this assumption by modeling the
independent and sparse behavior of the entities.

\subsection{Action segmentation}

Action segmentation is another line of works that share with us the
goal of finding temporal structure of activities in video. Notable
works include semi-Markov model \cite{shi2011human}, multi-class
SVM on spatial bag-of-words \cite{hoai2011joint}, and segmental RNNs
\cite{kong2015segmental}. More recently, convolutional neural networks
(CNNs) were more dominant as backbone for action segmentation methods
\cite{lea2017temporal,lea2016segmental,lei2018temporal}. For instance,
Farha and Gall \cite{farha2019ms-tcn} proposed a multi-stage CNN
(MS-TCN) with and a truncated MSE loss to handle over-segmentation.
Different to this line of works where activities are singularly modeled,
we explore the interactions between human and objects and consider
the relations between human sub-activities and object affordances.

\subsection{Sparse and asynchronous event modeling}

The sparsity and asynchronicity of events have been a modeling goal
of the signal processing community. Neil $\etal$ \cite{neil2016phased}
extended the long short-term memory (LSTM) cell \cite{hochreiter1997long}
formulation with a ``time'' gate. This gate introduces ``open''
and ``close'' cycles and allows for sparsity in the state updates.
Similarly, Campos $\etal$ \cite{campos2018skip} introduced sparsity
into the updates of RNNs by learning binary decisions regularized
by a budget loss to skip redundant state updates. In contrast to these
works, we skip state updates not only to reduce the computational
complexity but also to match the semantics of the human activities.
Furthermore, $\modelshort$ is more advanced in fully using the dense
input signal for the sparse activity decisions.

In processing naturally sparse signals, Sekikawa $\etal$ \cite{sekikawa2019eventnet}
proposed EventNet for real-time asynchronous event streams from event-based
cameras. EventNet process events via a two-module architecture timed
by the input events and output predictions. Asynchronous data from
event-based cameras are also handled by an extended version \cite{messikommer2020event-based}
of the Submanifold Sparse Convolutions (SSC) \cite{graham2018semantic}.
This work extend the spatial sparsity modeling of SSC with localized
updates throughout the convolutional maps by keeping track of a rulebook
per layer. They key difference between this line of work and our formulation
is that we explore sparse information from dense signals instead of
assuming the signal is already sparse.

\section{Method\label{sec:method}}

\subsection{Problem formulation}

We are interested in learning the spatio-temporal structures of human-object
interactions (HOI) in videos. Previous works consider special cases
of a single human \cite{koppula2016anticipating,qi2018learning} or
two human hands interacting with multiple objects \cite{dreher2020learning}.
We approach this problem in the most generic formulation, where we
model an arbitrary number of humans and objects in a video. The problem
is defined on a video of $T$ frames with $N$ entities of either
human or object class. These entities are detected and their features
are extracted from the video. The $e^{th}$ entity is represented
by a temporal sequence of frame-level features $X^{e}=\{x_{t}^{e}\}_{t=1,...,T}$
together with a class label $c^{e}$. In human-object interaction,
this label holds the value of either \emph{human} or \emph{object}.

A HOI recognition problem is defined to use input $\left\{ X^{e},c^{e}\right\} _{e=1,..N}$
to generate a \emph{temporal segmentation} for each entity included.
For the $e^{th}$ entity, the segmentation is of the form $S^{e}=\left\{ s_{1}^{e},s_{2}^{e},...,s_{n_{e}}^{e}\right\} $
in which each member segment is represented by its start time and
end time (which is the start time of next segment) $s_{k}^{e}=\left[t_{k}^{e},t_{k+1}^{e}\right]$.
The output also includes the prediction of segment labels $y^{e}=\left\{ y_{1}^{e},\ldots,y_{n_{e}}^{e}\right\} $
which effectively are sub-activity labels for humans and affordance
labels for objects.

For a human entity, a segment label is the name of a sub-activity,
and for an object entity it is the name of an affordance. These labels
are interlinked with each other; for example, a sub-activity ``drinking''
of a human usually overlaps with affordance ``drinkable'' of a cup.
However, they do not need to be perfectly aligned. An object can remain
in the same state during all human activities not involving it. Modeling
these sparse and asynchronous relations is the goal of this work.

\subsection{Asynchronous-Sparse Interaction Graph Network}

\begin{figure}

\centering{}\includegraphics[width=0.65\columnwidth]{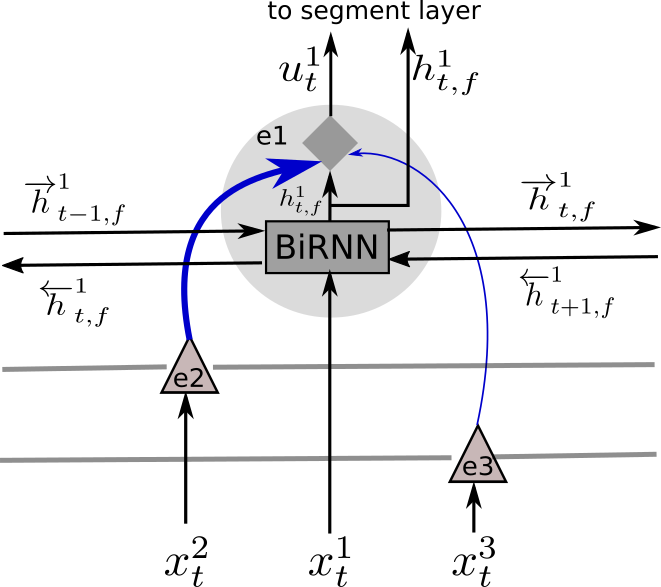}\caption{\emph{Frame-level node} (only enlarged and detailed for human node
e1 - the circle)\emph{ }with the BiRNN unit (rectangle) and\emph{
the segment boundary detector} (diamond shape). The detector considers
the current state from the recurrent unit and messages from neighboring
nodes (blue curved arrows) weighted by an attention mechanism (thickness
of arrows). It then makes a decision $u_{t}^{e}$ on whether frame
$t$ is the final frame of a segment for each entity. If it is a positive
signal ($u_{t}^{e}=1$), the summarized context $h_{t,f}^{1}$ is
sent up to the segment-level node to predict the label of finalized
segment and start a new one.\label{fig:State-change-detector}}
\end{figure}

We aim at learning the temporal segmentation and label of sparse events
associated with asynchronous entities in a video. To this end, we
design a two-layer asynchronous recurrent graph network named $\model$
($\modelshort$). $\modelshort$ is specialized in modeling each entity
in a video (human or object) with two spatio-temporal graphs, one
at frame-level and one at segment-level (Figure~\ref{fig:model-architecture}).
The frame-level graph nodes process video frames and update their
states at every time step, whereas the segment-level nodes update
sparsely - only when signaled by the frame-level partner to do so.
Each entity decides its own pace asynchronously in the consideration
to its neighbors. 

\subsection{Segmenting the entity life\label{subsec:Segmenting-the-entity}}

The primary task for $\modelshort$ is to learn the temporal segmentation
of every entity in a video. This translates into making a binary decision
at each time step of whether a current segment ends and a new segment
starts or not. The segment change of a sub-activity or affordance
depends on the internal state of the entity in question and its relation
with its neighbors. For example, a human that gets close to a cup
makes it a \emph{drinkable} object. This insight is realized into
the design of the frame-level layer of $\modelshort$ (Fig.~\ref{fig:State-change-detector}).

This network takes as input $\left\{ X^{e},c^{e}\right\} _{e=1,..N}$
and builds a spatio-temporal graph. Spatial edges represent interactions
between entities, and temporal edges connect instances of the same
entity throughout time and represent the internal progression of such
an entity. A temporal edge is implemented as a Bidirectional RNN (BiRNN)
and generates the hidden state of the $e$-th entity at the $t$-th
frame by:
\begin{align}
h_{t,f}^{e} & =\mathrm{BiRNN}_{f}(x_{t}^{e},\overrightarrow{h}_{t-1,f}^{e},\overleftarrow{h}_{t+1,f}^{e}),
\end{align}
where $\overrightarrow{h}_{t-1,f}^{e}$ and $\overleftarrow{h}_{t+1,f}^{e}$
are input forward and backward RNN states, and $h_{t,f}^{e}$ is the
concatenated output of the two RNNs: $h_{t,f}^{e}=\left[\overrightarrow{h}_{t,f}^{e},\overleftarrow{h}_{t,f}^{e}\right]$.

The spatial edges connect different entities at the same time step
and reflect the dynamic relations of neighboring entities. It is implemented
by pair-wise messaging between entities. We separate the two types
of spatial messages: (1) \emph{intra-class messages} from entities
of the same class and (2) \emph{inter-class messages }from entities
of different classes. This separation is important because the nature
of the relations are different. For example, the collaboration between
two human entities must be modeled differently than the effect of
an object on a human.

The frame-level inter-class message to entity $e$ at time $t$ is
calculated by:
\begin{equation}
m_{t,f}^{\textrm{inter\ensuremath{\rightarrow e}}}=\mathrm{Att}\left([x_{t}^{e};h_{t,f}^{e}],\left\{ [x_{t}^{k};h_{t,f}^{k}]\right\} {}_{c^{k}\neq c^{e}}\right).\label{eq:frame-level-inter-entity-message}
\end{equation}

Here, $\textrm{Att}$ is the attention operator that calculates the
weighted average of the contributions from the neighboring entities.
In ASSIGN, it is implemented by a variant of scaled dot-product attention
\cite{vaswani2017attention} with identical keys and values:
\begin{align}
\mathrm{Att}\left(q,\left\{ v_{i}\right\} _{i=1,...,n}\right) & =\sum_{i=1}^{n}\mathrm{softmax}\left(\frac{q^{T}v_{i}}{\sqrt{d}}\right)v_{i},\label{eq:att}
\end{align}
where $q$ is a query vector, $\left\{ v_{i}\right\} $ is a set of
keys/values vectors of size $n$ and, and $d$ is the feature dimension. 

Effectively, this operation combines the hidden states $h_{t,f}^{*}$
and inputs $x_{t}^{*}$ of the entities and use them as both keys/values
and queries in weighing the relevance of the interacted neighboring
nodes (the blue arrows in Fig.~\ref{fig:State-change-detector}).

Similarly, the intra-class message is calculated on the set of entities
from the same class:
\begin{equation}
m_{t,f}^{\textrm{intra\ensuremath{\rightarrow e}}}=\mathrm{Att}\left(\left[x_{t}^{e};h_{t,f}^{e}\right],\left\{ \left[x_{t}^{k};h_{t,f}^{k}\right]\right\} {}_{k\neq e,c^{k}=c^{e}}\right).\label{eq:frame-level-intra-entity-message}
\end{equation}

These spatial edges resemble a graph attention network \cite{velickovic2018graph}
except that they dynamically evolve through time.

Finally, we gather the current temporal recurrent state together with
the spatial relation messages to make the segmenting decision. This
is done by the \emph{segment boundary detector} (diamond shape in
Fig.~\ref{fig:State-change-detector}). It contains an MLP $\gamma$
and a differentiable discrete valued estimator using the Gumbel-Softmax
(GSM) operator \cite{jang2016categorical,maddison2016concrete}:
\begin{align}
u_{t}^{e} & =\textrm{GSM}\left\{ \gamma\left(\left[x_{t}^{e};h_{t,f}^{e};m_{t,f}^{\textrm{intra}};m_{t,f}^{\textrm{inter}}\right]\right)\right\} .\label{eq:change_det}
\end{align}
The binary output $u_{t}^{e}=1$ indicates that $t$ is the last frame
of the current segment for the $e$-th entity and $u_{t}^{e}=0$ otherwise.
This segmenting signal controls the behavior of the segment-level
nodes, which we describe next.

\subsection{Labeling the learned segments\label{subsec:Labeling-the-learned}}

The segment-layer of $\modelshort$ manages the spatio-temporal dynamics
of the segments whose boundaries are provided by the frame-layer via
the segmenting signal $u_{t}^{e}$ and frame-level state $h_{t,f}^{e}$.
This layer is also modeled as a spatio-temporal graph similarly to
the frame-layer with BiRNN for temporal edges and attentional message
passing for spatial connections.

The key specialty of this graph layer is that its operations are not
dense and regular as in the frame layer. Each entity can either update
or copy its state, depending on the provided signal $u_{t}^{e}.$
This adaptive operation constitutes the asynchronous and sparse behavior
of $\modelshort$.

At each time step $t$, if $u_{t}^{e}=1$, the node gathers information
from the context and update its state using its recurrent operator
(blue arrows in the upper half of Fig.\ref{fig:model-architecture}).
This includes the segment-level inter-class message
\begin{align}
m_{t,s}^{\intersymbol\rightarrow e} & =\mathrm{Att}\left(h_{t-1,s}^{e},\left\{ h_{t-1,s}^{k}\right\} {}_{c^{k}\neq c^{e}}\right),\label{eq:segment-level-inter-entity-message}
\end{align}
and intra-class message
\begin{align}
m_{t,s}^{\intrasymbol\rightarrow e} & =\mathrm{Att}\left(h_{t-1,s}^{e},\left\{ h_{t-1,s}^{k}\right\} {}_{k\neq e,c^{k}=c^{e}}\right),
\end{align}
where Att is defined in Eq.~\ref{eq:att}. These messages are calculated
similarly to frame-level counter parts in Eqs. \ref{eq:frame-level-inter-entity-message}
and \ref{eq:frame-level-intra-entity-message}. The main distinction
is that they are calculated sparsely, only when needed.

We combine these segment-level messages with the frame-level state
$h_{t,f}^{e}$ and messages $m_{t,f}^{\intersymbol\rightarrow e}$,
$m_{t,f}^{\intrasymbol\rightarrow e}$ previously calculated by the
frame layer to form the segment-level feature $z_{t}^{e}$:
\begin{align}
z_{t}^{e} & =\left[h_{t,f}^{e};m_{t,f}^{\intersymbol\rightarrow e};m_{t,f}^{\intrasymbol\rightarrow e};m_{t,s}^{\intersymbol\rightarrow e};m_{t,s}^{\intrasymbol\rightarrow e}\right].
\end{align}
This input is fed to the segment-level BiRNN units ($\textrm{BiRNN}_{s}$)
to update their states:
\begin{equation}
h_{t,s}^{e}=\mathrm{BiRNN}_{s}(z_{t}^{e},\overrightarrow{h}_{t-1,s}^{e},\overleftarrow{h}_{t+1,s}^{e}).
\end{equation}
The updated state is then used to recognize the label of the finished
segment:
\begin{align}
\hat{y}_{t}^{e} & =\textrm{Softmax}\left(\sigma(h_{t,s}^{e})\right),\label{eq:label-recognition}
\end{align}
where $\sigma$ is an MLP and the Softmax is done on appropriate label
sets, either human sub-activities or object affordances.

In the other case where $u_{t}^{e}=0$ , the node skips a BiRNN update
and maintain its current state. This contextualized skipping not only
creates sparsity in the state updates but also in the interactions.
The inward messages are skipped while outward messages to other updating
neighbors can still happen.

This is a better reflection of the world where at-rest entities ($\eg$
objects far from humans) can avoid unnecessary state updates and over-segmented
predictions. It also prevents the short-term memory of the RNNs from
fading quickly. Furthermore, operating at the segment-level separate
semantic progress (activity) from raw signals (frames) so is more
robust to varied video sampling rates. The sophisticated architecture
of $\modelshort$ requires a customized training procedure, which
we describe in the next section.

\subsection{Model training\label{subsec:model-training}}

ASSIGN is effectively a multi-task learning framework where segmentation
and labeling tasks are trained together in an end-to-end fashion.
It is therefore trained by an ensemble of two losses for the two tasks.

For segmentation, we minimize the binary cross-entropy between a smoothed
version of the ground-truth segmentation and the soft output of the
boundary detector in Eq.~\ref{eq:change_det} 
\begin{equation}
\mathcal{L_{\textrm{Seg}}}=\frac{1}{T}\sum_{t=1}^{T}\left[\frac{1}{N}\sum_{e=1}^{N}\mathrm{BCE}(\hat{\tilde{u}}_{t}^{e},\tilde{u}_{t}^{e})\right],\label{eq:loss_seg}
\end{equation}
where $\hat{\tilde{u}}_{t}^{e}$ is the real value of $u_{t}^{e}$
before binary thresholding and $\tilde{u}_{t}^{e}$ is the smoothed
version (with Gaussian filter of $\sigma=4$) of the binary pulse
ground-truth segmentation.

For labeling, we minimize the negative log-likelihood of the predicted
sub-activities and affordance labels:
\begin{align}
\mathcal{L}_{\textrm{Label}} & =\frac{1}{T}\sum_{t=1}^{T}\left[\frac{1}{N}\sum_{n=1}^{N}\mathrm{NLL}(\hat{y}_{t}^{n},y_{t}^{n})\right].
\end{align}
Even though the label is predicted per segment, this loss is calculated
per frame so that long segments contribute more than short ones.

The overall loss is the weighted sum of the two losses
\begin{equation}
\mathcal{L}=\mathcal{L}_{\textrm{Label}}+\lambda\mathcal{L}_{\textrm{Seg}},
\end{equation}
where $\lambda$ is a tunable parameter.

While the sparsity of $\modelshort$ operations provide strong advantage
in authentic modeling, it creates a subtle obstacle in training where
informative gradients from the labeling loss in the segment-layer
rarely reaches the frame-layer. This phenomenon makes the frame-layer
struggle to have enough training signal to learn adequately. To overcome
this issue, we use a two-stage training procedure. In stage 1, we
switch off $\mathcal{L_{\textrm{Seg}}}$ and set $u_{t}^{e}:=1$ everywhere
so that the frame-layer receives a constant stream of directive signals.
In stage 2, we turn on the full model and continue to train on top
of parameters learned in stage 1. We observed in our experiments that
this two-stage training leads to faster convergence and better final
results.

\subsection{Implementation details}

For entity features, we use 2048-dimensional ROI pooling features
extracted from the 2D bounding boxes of humans and objects in the
video detected by a Faster R-CNN \cite{ren2017faster} module pre-trained
\cite{anderson2018bottom-up} on the Visual Genome dataset \cite{krishna2017visual}.
We optimize model parameters using the ADAM optimizer \cite{kingma2014adam},
with a learning rate of $10^{-3}$. All recurrent networks are built
with Gated Recurrent Units (GRU) \cite{cho2014learning}. The videos
are resampled to a uniform 10 FPS frame rate before feeding into the
frame-level network. For model validation, we utilize 10\% of the
training data as validation data, and select the model with the lowest
validation loss during training.

\section{Experiments}

\begin{table*}
\caption{\emph{Joint segmentation and label recognition} task with no pre-segmentation.
Performance on the CAD-120 dataset. \label{tab:recognition-without-ground-truth-segmentation-cad120}}

\centering{}\resizebox{.80\textwidth}{!}{
\begin{tabular}{ccccccccc}
\toprule 
\multirow{2}{*}{Model} &  & \multicolumn{3}{c}{Sub-activity} &  & \multicolumn{3}{c}{Object Affordance}\tabularnewline
 &  & $\mathrm{F}_{1}@0.10$ & $\mathrm{F}_{1}@0.25$ & $\mathrm{F}_{1}@0.50$ &  & $\mathrm{F}_{1}@0.10$ & $\mathrm{F}_{1}@0.25$ & $\mathrm{F}_{1}@0.50$\tabularnewline
\cmidrule{1-1} \cmidrule{3-5} \cmidrule{4-5} \cmidrule{5-5} \cmidrule{7-9} \cmidrule{8-9} \cmidrule{9-9} 
rCRF \cite{sener2015rCRF} &  & 65.6 $\pm$ 3.2 & 61.5 $\pm$ 4.1 & 47.1 $\pm$ 4.3 &  & 72.1 $\pm$ 2.5 & 69.1 $\pm$ 3.3 & 57.0 $\pm$ 3.5\tabularnewline
Independent BiRNN &  & 70.2 $\pm$ 5.5 & 64.1 $\pm$ 5.3 & 48.9 $\pm$ 6.8 &  & 84.6 $\pm$ 2.1 & 81.5 $\pm$ 2.7 & 71.4 $\pm$ 4.9\tabularnewline
ATCRF \cite{koppula2016anticipating} &  & 72.0 $\pm$ 2.8 & 68.9 $\pm$ 3.6 & 53.5 $\pm$ 4.3 &  & 79.9 $\pm$ 3.1 & 77.0 $\pm$ 4.1 & 63.3 $\pm$ 4.9\tabularnewline
Relational BiRNN &  & 79.2 $\pm$ 2.5 & 75.2 $\pm$ 3.5 & 62.5 $\pm$ 5.5 &  & 82.3 $\pm$ 2.3 & 78.5 $\pm$ 2.7 & 68.9 $\pm$ 4.9\tabularnewline
\cmidrule{1-1} \cmidrule{3-5} \cmidrule{4-5} \cmidrule{5-5} \cmidrule{7-9} \cmidrule{8-9} \cmidrule{9-9} 
$\modelshort$ &  & \textbf{88.0} $\pm$ 1.8 & \textbf{84.8} $\pm$ 3.0 & \textbf{73.8} $\pm$ 5.8 &  & \textbf{92.0} $\pm$ 1.1 & \textbf{90.2} $\pm$ 1.8 & \textbf{82.4} $\pm$ 3.5\tabularnewline
\bottomrule
\end{tabular}}
\end{table*}

\begin{table}
\caption{\emph{Label recognition only} task with ground-truth segmentation.
Performance on the CAD-120 dataset. Unreported results are marked
as \textquotedblleft -\textquotedblright .\label{tab:recognition-with-ground-truth-segmentation-cad120}}

\centering{}\resizebox{\columnwidth}{!}{
\begin{tabular}{ccccccc}
\toprule 
\multirow{2}{*}{Model} &  & \multicolumn{2}{c}{Sub-activity $\mathrm{F}_{1}$ (\%)} &  & \multicolumn{2}{c}{Object Affordance $\mathrm{F}_{1}$ (\%)}\tabularnewline
 &  & Micro  & Macro  &  & Micro  & Macro \tabularnewline
\cmidrule{1-1} \cmidrule{3-4} \cmidrule{4-4} \cmidrule{6-7} \cmidrule{7-7} 
GPNN \cite{qi2018learning} &  & 76.6  & 72.7  &  & 74.6  & 54.1 \tabularnewline
S-RNN (multi-task) \cite{jain2016structural-rnn} &  & 82.4 & - &  & 91.1 & -\tabularnewline
KGS \cite{koppula2013learning} &  & 86.0  & 80.4  &  & 91.8  & 81.5 \tabularnewline
Latent Linear-CRF \cite{hu2014learning} &  & 87.0  & 86.0 &  & - & -\tabularnewline
ATCRF \cite{koppula2016anticipating} &  & 89.3  & 86.4 &  & 93.9  & 85.7 \tabularnewline
STGCN \cite{ghosh2020stacked} &  & - & 87.2 &  & - & -\tabularnewline
\cmidrule{1-1} \cmidrule{3-4} \cmidrule{4-4} \cmidrule{6-7} \cmidrule{7-7} 
$\modelshort$ &  & \textbf{89.9}  & \textbf{87.8}  &  & \textbf{95.9}  & \textbf{91.9} \tabularnewline
\bottomrule
\end{tabular}}
\end{table}

\begin{table}
\caption{Joint segmentation and label recognition task \emph{with multiple
human entities}. Performance on Bimanual Actions dataset.\label{tab:recognition-without-ground-truth-segmentation-bimanual}}

\centering{}\resizebox{.90\columnwidth}{!}{
\begin{tabular}{ccccc}
\toprule 
\multirow{2}{*}{Model} &  & \multicolumn{3}{c}{Sub-activity}\tabularnewline
\cmidrule{3-5} \cmidrule{4-5} \cmidrule{5-5} 
 &  & $\mathrm{F}_{1}@0.10$ & $\mathrm{F}_{1}@0.25$ & $\mathrm{F}_{1}@0.50$\tabularnewline
\cmidrule{1-1} \cmidrule{3-5} \cmidrule{4-5} \cmidrule{5-5} 
Dreher $\etal$ \cite{dreher2020learning} &  & 40.6 $\pm$ 7.2 & 34.8 $\pm$ 7.1 & 22.2 $\pm$ 5.7\tabularnewline
Independent BiRNN &  & 74.8 $\pm$ 7.0 & 72.0 $\pm$ 7.0 & 61.8 $\pm$ 7.3\tabularnewline
Relational BiRNN &  & 77.7 $\pm$ 3.9 & 75.0 $\pm$ 4.2 & 64.8 $\pm$ 5.3\tabularnewline
\cmidrule{1-1} \cmidrule{3-5} \cmidrule{4-5} \cmidrule{5-5} 
$\modelshort$ &  & \textbf{84.0} $\pm$ 2.0 & \textbf{81.2} $\pm$ 2.0 & \textbf{68.5} $\pm$ 3.3\tabularnewline
\bottomrule
\end{tabular}}
\end{table}

\subsection{Datasets}

We evaluate $\modelshort$ on the CAD-120 \cite{koppula2013learning}
and on the Bimanual Actions \cite{dreher2020learning} datasets. CAD-120
is the most popular dataset for HOI recognition. It contains 120 RGB-D
videos of 4 subjects executing 10 different activities, each activity
repeated 3 times. Each video is composed of a single person interacting
with 1-5 objects. Sub-activities of the person the affordances of
every object are annotated with 10 sub-activity and 12 object affordance
labels. 

We also experiment on the Bimanual Actions - the first HOI dataset
of activities featuring a subject using both hands to interact with
objects simultaneously ($\eg$ left hand holds a nail while right
hand hits the nail). It has 540 RGB-D videos of 6 subjects conducting
9 different tasks, each task repeated 10 times. The actions of each
hand are annotated frame-wise as one of 14 possible actions. For both
datasets, we only use the RGB channels to extract frame features.

\subsection{Experimental settings\label{subsec:experimental-settings}}

We evaluate $\modelshort$ on two tasks: joint segmentation and label
recognition, and label recognition with known segmentation. The first
task requires models to segment the time line for each entity in a
video and label those segments. The second task is a special case
of the first one where the ground-truth segmentation is known and
models only need to label the provided segments.

To evaluate how well $\modelshort$ generalizes to unseen subjects,
we do leave-one-subject out cross-validation on both datasets. Previous
works concentrated on recognition of labels and usually report frame-level
$\mathrm{F}_{1}$ scores. However, these metrics are not optimal for
tasks involving segmentation because a method might heavily over-
or under-segment a video and still attain reasonable frame-level scores.
To amend that, we use the $\mathrm{F}_{1}@k$ metric \cite{lea2017temporal}
for the commonly used values of $k=0.10,0.25,$ and $0.50$. The $\mathrm{F}_{1}@k$
metric considers a predicted segment correct if its IoU with the ground-truth
segment is at least $k$. Wrong predictions and missed ground-truth
segments count as false positives and false negatives, respectively.
The $\mathrm{F}_{1}@k$ is a superior choice over frame-based metrics
for joint segmentation and labeling problems, and widely used in previous
segmentation works \cite{farha2019ms-tcn,lea2017temporal,morais2020learning}.
Note that for the label recognition with known segmentation task,
$\mathrm{F}_{1}@k$ is constant for any $k$ and reduced to segment-level
micro and macro $\mathrm{F}_{1}$ scores. We report these metrics,
in line with other reported results in the literature.

\subsection{Quantitative results\label{subsec:quantitative-results}}

\subsubsection{Joint segmentation and label recognition\label{subsec:Joint-task}}

In this main experiment, we compare the performance of $\modelshort$
with related state-of-the-art and two BiRNN-based baselines on the
joint segmentation and label recognition task on the CAD-120 dataset.
For this task, the input must be raw video features, with no trace
of preproduced segmentation.

Two of the previous works fully qualify for this task: ATCRF \cite{koppula2016anticipating}
and rCRF \cite{sener2015rCRF}. Other major related works used the
preproduced segmentation information in either explicit or implicit
ways. Stochastic grammar \cite{qi2017predicting} used the statistics
of segmentation from test portion in training. Earley tree parser
\cite{qi2018generalized} repeats the preproduced segment-level features
as frame-level features hence implicitly acknowledging the true segment
boundaries. More concrete details about these uses are included in
the Appendix~\ref{app:experimental-qualification}.

The baselines are two variations of BiRNN GRU: The \emph{Independent
BiRNN} considers each entity only by its temporal edges, and the \emph{Relational
BiRNN} adds dense spatial connections across entities. (Further details
in Appendix~\ref{app:baselines}.)

We present the $\mathrm{F}_{1}@k$ results in Table~\ref{tab:recognition-without-ground-truth-segmentation-cad120},
and in frame-level $\mathrm{F}_{1}$ in Appendix~\ref{app:results-micro-macro}. $\modelshort$
outperforms both the state-of-the-art and baselines in every configuration
of the $\mathrm{F}_{1}@k$ measure, and for both human sub-activities
and object affordances.

This result showcases the advantages of doing segmenting and labeling
jointly. Other methods employ separate segmentation and labeling steps,
and generate their final result by voting on many different segmentation
options. This strategy has a weakness of increasing the over-segmentation
when voters disagree and inability to correct if they make the same
mistakes. 

The BiRNN baselines make frame-wise predictions and lack relational
modeling, thus they do not fully leverage the human-object interactions.
Despite being simpler, the Independent BiRNN is better than the Relational
BiRNN in object affordances. This can be explained by the infrequent
changes of object affordances that were mistaken by the presence of
dense messages from the human nodes. In contrast, $\modelshort$ allows
sparse messaging and effectively overcomes these problems.

\subsubsection{Label recognition only}

To examine the sole capability of predicting labels and to match with
the task done by more previous works, we setup a simpler experiment
where the true segmentation is provided to all methods. This skips
the segmentation functionality of $\modelshort$ and put it in fair
comparison with all previous works in their common experimental protocol. 

Table~\ref{tab:recognition-with-ground-truth-segmentation-cad120}
presents the micro and macro $\mathrm{F}_{1}$ accuracy on CAD-120
dataset. In both metrics and for both sub-activity and object affordance,
$\modelshort$ outperforms all other methods. This further demonstrates
that our modeling entities with asynchronous and sparse interactions
is a more correct way to label the segments, agnostic to the segmentation
quality.

\subsubsection{Multiple human entities}

The generic formulation of $\modelshort$ makes it easily applied
to a wide range of scenarios. We test this capability by trialing
it in the case where multiple human entities jointly do a task. This
experiment is done on the Bimanual Actions dataset, which contains
activities of two hands of the same person interacting with many objects
in a shared activity. We compare $\modelshort$ to the BiRNN baselines
(Sec. \ref{subsec:Joint-task}) and the method of Dreher $\etal$
\cite{dreher2020learning}, which is the only previous work proposed
for this multi-human setting.

Table~\ref{tab:recognition-without-ground-truth-segmentation-bimanual}
compares the performance of these methods on the joint segmentation
and labeling task. The method of Dreher $\etal$ \cite{dreher2020learning}
has the weakest performance. This can be attributed to their over-simplistic
graph network, which ignores the interactions between the two hands
and does not take long-term temporal context into account. The BiRNN
baselines improve over Dreher $\etal$ \cite{dreher2020learning}
by considering longer temporal context but fall short in reaching
high accuracy for lacking modeling human-human interactions. $\modelshort$
makes major improvements over these methods by incorporating cross-hand
spatial interaction and asynchronous long-term temporal context. This
good performance is also attributed to the segment-level label decisions
in contrast to frame-based decisions of the baselines.

Through the quantitative experiments, it is clear that joint structure-content
exploration with consideration to the temporal object lives are key
features of $\modelshort$ that makes it excel in recognition performance.
The qualitative analysis of outputs and internal operation of the
model are examined in the next section.

\subsection{Qualitative analysis\label{subsec:qualitative-results}}

\begin{figure}
\centering{}\includegraphics[width=1\columnwidth]{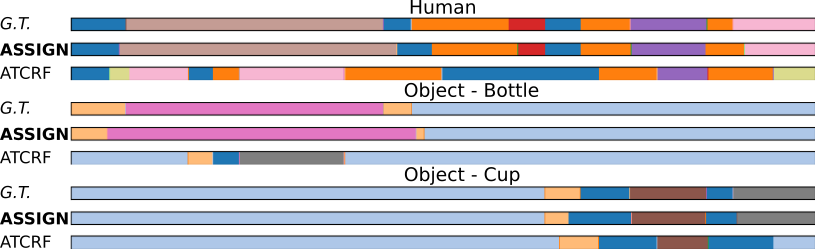}\caption{Segmentation and labeling results of the top performing $\protect\modelshort$
and ATCRF methods compared to ground-truth on the CAD-120 dataset
for a \emph{taking medicine} activity. In this example, ATCRF over-segments
the long \emph{opening} ($\protect\lightbrownbox$) sub-activity for
the human. Because objects are synchronized with human in ATCRF, these
over-segmentations created a domino effect that leads to the incoherent
structure of the timeline of Object~-~Bottle. In contrast, $\protect\modelshort$
allow asynchronous state changes of human and objects thus avoided
this type of mistakes. Legends: Sub-activities - $\protect\darkbluebox$
\emph{reaching, }$\protect\lightbrownbox$ \emph{opening, }$\protect\orangebox$
\emph{moving, $\protect\redbox$ eating}, $\protect\purplebox$ \emph{drinking},
$\protect\lightpinkbox$ \emph{placing}, and $\protect\palemustardbox$
\emph{null}; Affordances - $\protect\paleorangebox$ \emph{reachable},
$\protect\darkpinkbox$ \emph{openable,} $\protect\palebluebox$ \emph{stationary},
$\protect\darkbluebox$ \emph{movable}, $\protect\darkbrownbox$ \emph{drinkable},
and $\protect\darkgraybox$ \emph{placeable}.\label{fig:qualitative-results-cad120}}
\end{figure}
\begin{figure}
\includegraphics[width=1\columnwidth]{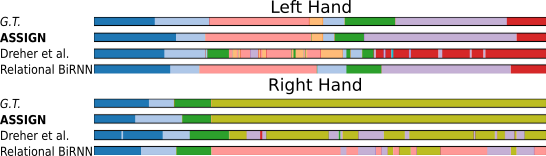}\caption{Segmentation and labeling results on the Bimanual dataset for a \emph{cooking}
task. In this example, Dreher $\etal$ \cite{dreher2020learning}
creates many spurious short segments due to their model's limited
temporal context. The Relational BiRNN baseline improves on short
activities, but fails to handle longer events such as the long \emph{stir
}($\protect\palegreenbox$) action because the recurrent memory forgets
quickly. On the other hand, $\protect\modelshort$ handles long actions
well by appropriately skipping redundant updates. Legend: $\protect\darkbluebox$
\emph{idle}, $\protect\palebluebox$ \emph{approach}, $\protect\greenbox$
\emph{lift}, $\protect\palegreenbox$ \emph{stir}, \emph{$\protect\pinksalmonbox$
hold,} $\protect\paleorangebox$ \emph{retreat}, $\protect\lightpurplebox$
\emph{pour}, and $\protect\redbox$ \emph{place}.\label{fig:qualitative-results-bimanual}}

\end{figure}

We compare the detail outputs of $\modelshort$ and related methods
on examples from CAD-120 and Bimanual Actions datasets. Figure~\ref{fig:qualitative-results-cad120}
shows an example in CAD-120 where ATCRF over-segments human sub-activities.
Also, because segments are synchronized between entities, these errors
spread to objects and hurt the accuracy on affordance recognition.
$\modelshort$ on the other hand successfully overcomes the over-segmentation
and the error propagation by supporting sparse and asynchronous processing.

Figure~\ref{fig:qualitative-results-bimanual} shows an example of
a \emph{cooking} task on the Bimanual dataset. Dreher $\etal$ \cite{dreher2020learning}
take limited temporal context into account and creates many short
segments. The Relational BiRNN improves on short segments but fails
to handle long ones ($\eg$ long \emph{stir} action of the right hand).
$\modelshort$, with more advanced modeling, is well equipped to reliably
handle both short- and long-term interactions.

\begin{figure}
\centering{}\includegraphics[width=1\columnwidth]{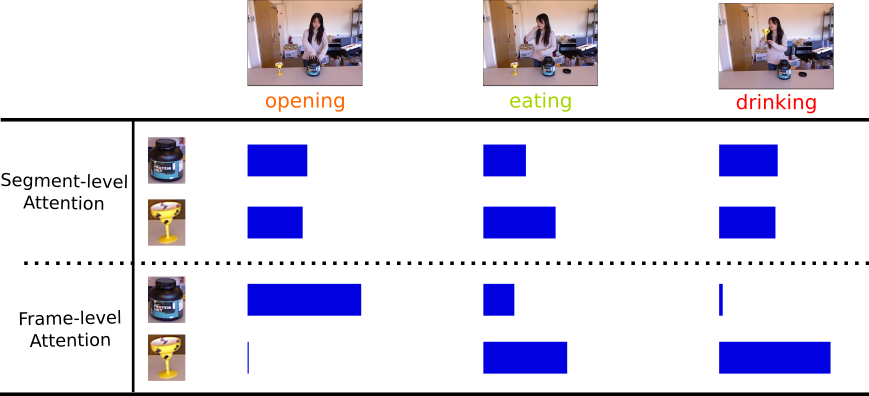}\caption{Attention scores of the messages from the objects to the human entity
at both layers. Sharp and strong attention on relevant objects are
used in frame-level to gather key information. More uniform attention
is found in the segment-level where overall consideration is made.
\label{fig:attention-scores-assign}}
\end{figure}

In Figure~\ref{fig:attention-scores-assign}, we analyze the attention
scores of the objects in relation to the human at both levels of $\modelshort$.
At the frame-level, the human entity is shown to pay sharper attention
to a specific object to make crisp decision on transitioning its activity.
At the segment level the attention is more uniform. This is reasonable
given the sparsity of the updates. At each sparse deciding point,
the human entity needs to consider multiple neighboring objects to
recognize the label of its sub-activity.

\subsection{Ablation study}

\begin{table}
\caption{Ablation study on the CAD-120 dataset.\label{tab:ablation-cad120}}

\centering{}\resizebox{\columnwidth}{!}{
\begin{tabular}{cccccccc}
\toprule 
 & \multirow{2}{*}{Model} &  & \multicolumn{2}{c}{Sub-activity} &  & \multicolumn{2}{c}{Object Affordance }\tabularnewline
 &  &  & $\mathrm{F}_{1}@0.10$ & $\mathrm{F}_{1}@0.50$ &  & $\mathrm{F}_{1}@0.10$ & $\mathrm{F}_{1}@0.50$\tabularnewline
\cmidrule{1-2} \cmidrule{2-2} \cmidrule{4-5} \cmidrule{5-5} \cmidrule{7-8} \cmidrule{8-8} 
1 & w/o message passing &  & 74.5 & 55.0 &  & 89.0 & 74.4\tabularnewline
2 & w/o segmentation loss &  & 84.3 & 69.9 &  & 89.2 & 78.6\tabularnewline
3 & w/ dense update &  & 85.5 & 70.3 &  & 90.6 & 79.8\tabularnewline
4 & w/o pre-training &  & 87.6 & 71.6 &  & 91.1 & 78.9\tabularnewline
\cmidrule{1-2} \cmidrule{2-2} \cmidrule{4-5} \cmidrule{5-5} \cmidrule{7-8} \cmidrule{8-8} 
5 & full $\modelshort$ model &  & \textbf{88.0} & \textbf{73.8} &  & \textbf{92.0} & \textbf{82.4}\tabularnewline
\bottomrule
\end{tabular}}
\end{table}

To understand the role of individual components of $\modelshort$,
we ablate several key modules and evaluate these variants on the CAD-120
dataset. Firstly, the spatial message passing has a crucial role in
modeling the entity interactions (row 1). On such graph, we join labeling
with segmenting tasks by adding the segmentation loss. This loss is
also essential for good joint results (row 2). On top of this joint
training scheme, $\modelshort$ is special by using asynchronous and
sparse interaction constraints. This innovation is proved to significantly
improve the robustness to a wide variety of activity structures (row
3 vs 5). Finally, the strategy of pre-training $\modelshort$ with
the dense model (see Section~\ref{subsec:model-training}) benefits
the learning process and supports the model to reach the highest performance
(row 4 vs row 5).

\section{Conclusions}

We designed $\modelshort$, a two-layer graph network that explores
the activity structure concurrently with predicting its content. ASSIGN
models human-object interaction more correctly than previous methods
by allowing the participating entities to have independent asynchronous
lives. The interactions in ASSIGN are sparse, therefore more robust
to varied segment lengths and activity progression.

These advantages resulted in superior performance in multiple datasets.
Moreover, this performance is consistently strong over larger variations
of scenarios than any other method. Deep analysis into ASSIGN's operation
shows that the strong performance comes from the new ability to deal
with over- or under-segmentation mistakes that previous models suffered
from. The generic capability of distilling structure from time series
showcases that $\modelshort$ can be readily applicable to other domains
and applications.

\bibliographystyle{ieee_fullname}
\bibliography{arxiv}

\appendix

\section{Qualification for \emph{joint segmentation and label recognition}
task\label{app:experimental-qualification}}

As mentioned in Section~\ref{subsec:Joint-task}, \cite{qi2017predicting,qi2018generalized}
implicitly or explicitly used information relevant to the ground-truth
segmentation during the training or testing of their models.

For the Stochastic Grammar method \cite{qi2017predicting}, the model
requires the computation of data statistics such as length of sub-activities
and object affordances, and these were computed from the whole dataset.
The experimental protocol is a leave-one-subject out cross-validation,
which requires such data-dependent statistics to be computed from
the training folds during each round of cross-validation. The code
relies on these pre-computed statistics to execute, and in the files
provided by the authors in a Github issue\footnote{https://github.com/SiyuanQi/grammar-activity-prediction/issues/2},
we can verify that the lengths of sub-activities and object affordances
provided are in relation to the full dataset.

For the Generalized Earley Parser method \cite{qi2018generalized},
the authors implicitly give segmentation information about the data
to their model by repeating the segment-level features provided by
Koppula $\etal$ \cite{koppula2013learning} as frame-level features.
We can confirm that by analyzing the function collate\_fn\_cad\footnote{https://github.com/SiyuanQi/generalized-earley-parser/blob/master/src/python/datasets/utils.py},
where the length information about the segments is used in lines 40-44
to assemble the frame-level features. This means that the segmentation
is implicitly input to their model, which makes their method not suitable
for this task.

\begin{figure}
\centering{}\includegraphics[width=1\columnwidth]{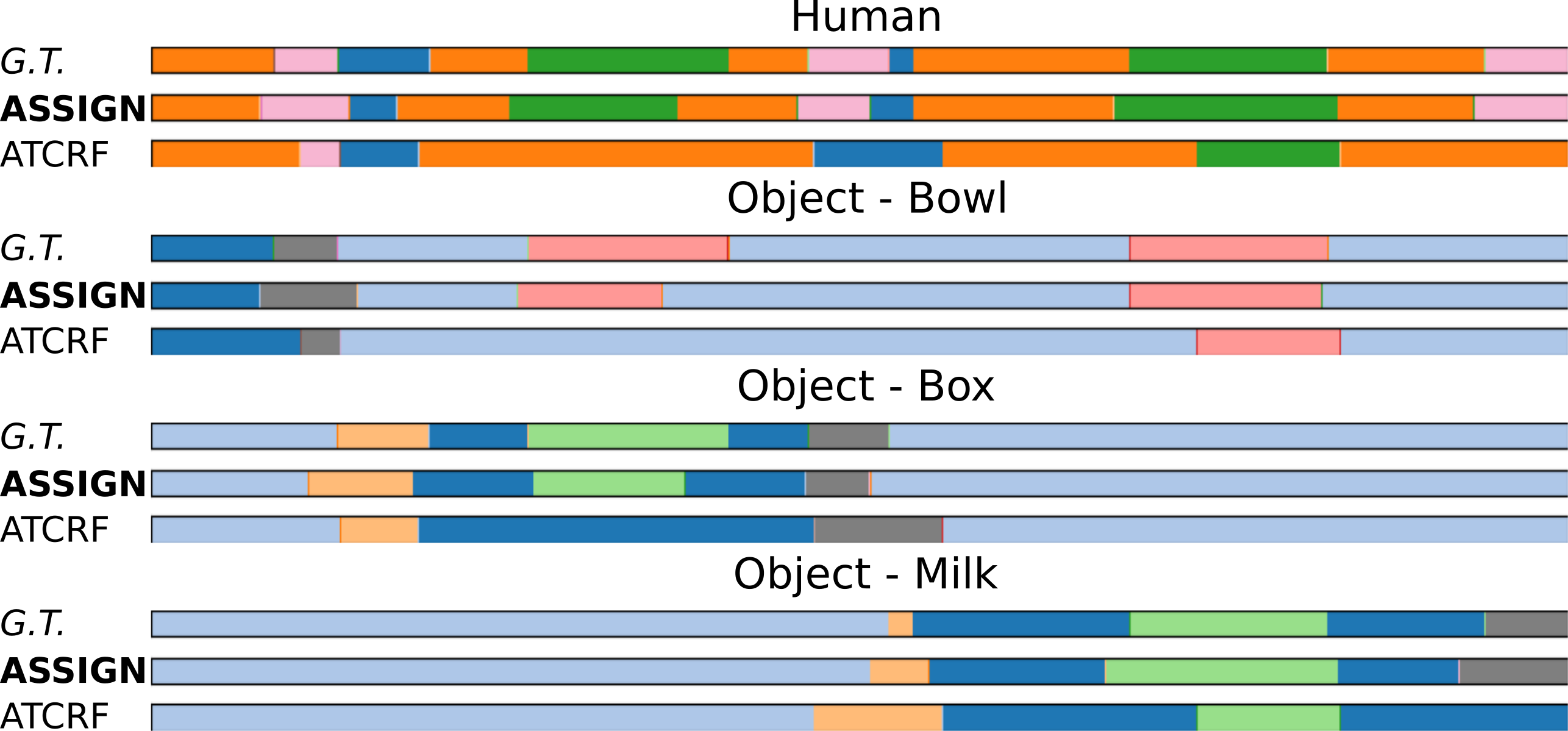}\caption{Segmentation and labeling results of ASSIGN and ATCRF on the CAD-120
dataset for a \emph{making cereal} activity. In this example, ATCRF
under-segments the human and the box by skipping a few segments and
merging adjacent labels. Sub-activities: $\protect\orangebox$ \emph{moving},
$\protect\lightpinkbox$ \emph{placing}, $\protect\darkbluebox$ \emph{reaching},
and $\protect\greenbox$ \emph{pouring}. Affordances: $\protect\darkbluebox$
\emph{movable}, $\protect\darkgraybox$ \emph{placeable}, $\protect\palebluebox$
\emph{stationary}, \emph{$\protect\lightredbox$} \emph{pour-to},
$\protect\paleorangebox$ \emph{reachable}, and $\protect\lightgreenbox$
\emph{pourable.}\label{fig:qualitative-cad120-making-cereal}}
\end{figure}

\begin{figure}
\centering{}\includegraphics[width=1\columnwidth]{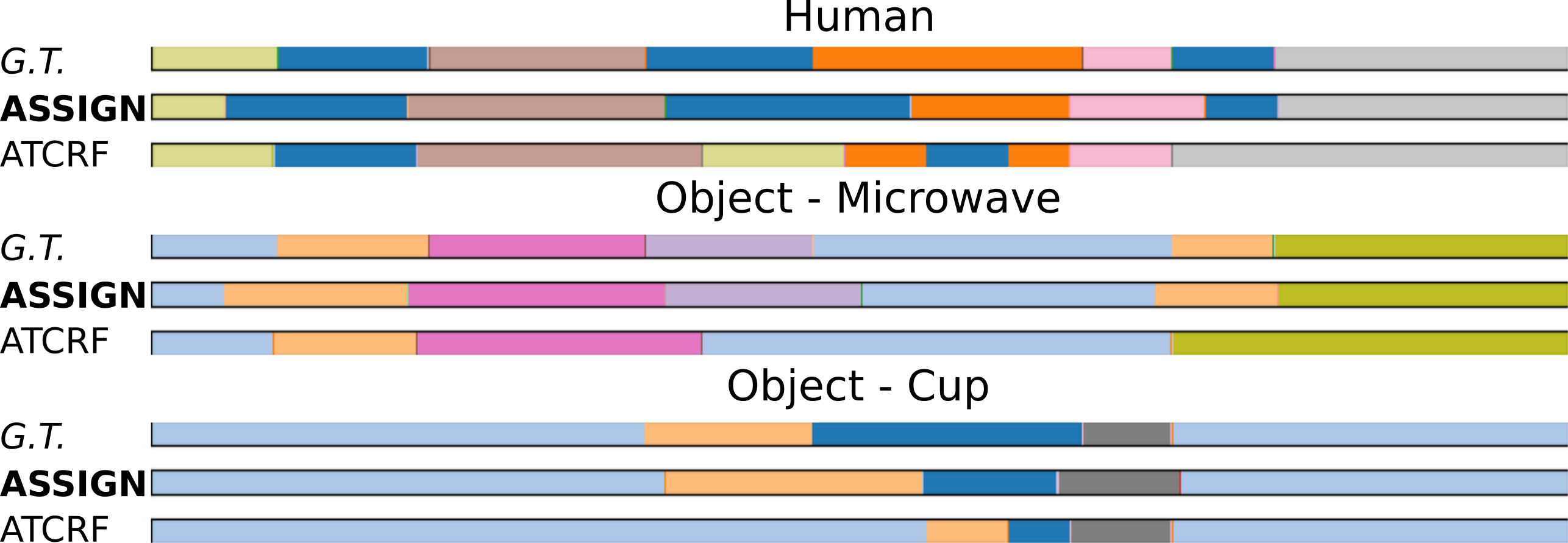}\caption{Segmentation and labeling results of ASSIGN and ATCRF on the CAD-120
dataset for a \emph{taking food} activity. In this example, ATCRF
over-segments the \emph{moving} ($\protect\orangebox$) sub-activity
and ignores the last \emph{reaching} ($\protect\darkbluebox$)\emph{.}
ASSIGN correctly segments and label the entities. Sub-activities:
$\protect\palemustardbox$ \emph{null}, $\protect\darkbluebox$ \emph{reaching},
$\protect\lightbrownbox$ \emph{opening}, $\protect\orangebox$ \emph{moving},
$\protect\lightpinkbox$ \emph{placing}, and $\protect\graybox$ \emph{closing}.
Affordances: $\protect\palebluebox$ \emph{stationary}, $\protect\paleorangebox$
\emph{reachable}, $\protect\darkpinkbox$ \emph{openable}, $\protect\lightpurplebox$
\emph{containable}, $\protect\palegreenbox$ \emph{closeable}, $\protect\darkbluebox$
\emph{movable}, and $\protect\darkgraybox$ \emph{placeable}.\label{fig:qualitative-cad120-taking-food}}
\end{figure}

\begin{figure*}
\centering{}\includegraphics[width=1\textwidth]{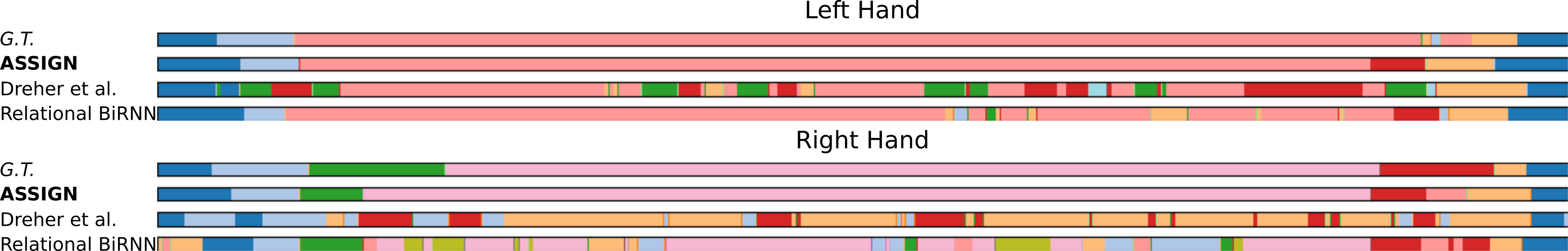}\caption{Segmentation and labeling results on the Bimanual Actions dataset
for a \emph{sawing} task. The main difficulty related methods have
is to handle long actions, such as the left hand \emph{hold} ($\protect\pinksalmonbox$).
Legend: $\protect\darkbluebox$ \emph{idle}, $\protect\palebluebox$
\emph{approach}, $\protect\pinksalmonbox$ \emph{hold}, $\protect\paleorangebox$
\emph{retreat}, $\protect\redbox$ \emph{place}, $\protect\greenbox$
\emph{lift}, and $\protect\lightpinkbox$ \emph{saw}.\label{fig:qualitative-bimanual-s1t9t7}}
\end{figure*}

\begin{figure*}
\centering{}\includegraphics[width=1\textwidth]{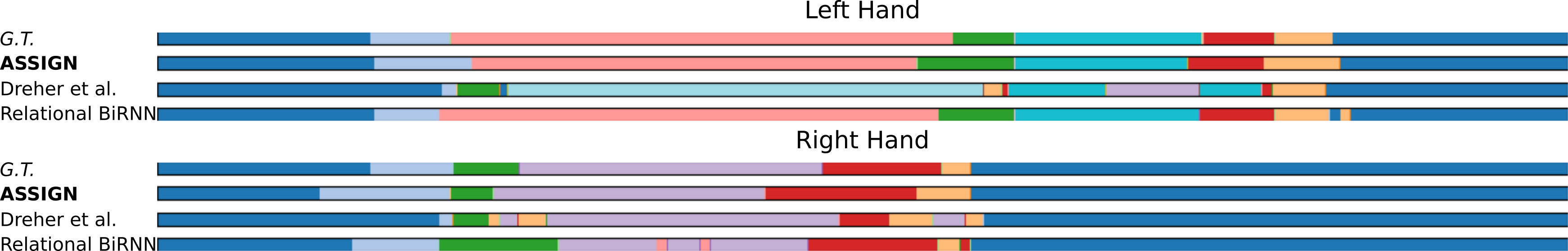}\caption{Segmentation and labeling results on the Bimanual Actions dataset
for a \emph{pouring} task. In this example, the Relational BiRNN has
results comparable to ASSIGN but still makes several over-segmentation
mistakes in both hands, such as the right hand \emph{pour }($\protect\lightpurplebox$).
Legend:$\protect\darkbluebox$ \emph{idle}, $\protect\palebluebox$
\emph{approach}, $\protect\pinksalmonbox$ \emph{hold}, $\protect\greenbox$
\emph{lift}, $\protect\turquoisebox$ \emph{drinking}, $\protect\redbox$
\emph{place}, $\protect\paleorangebox$ \emph{retreat}, and $\protect\lightpurplebox$
\emph{pour}.\label{fig:qualitative-bimanual-s6t3t7}}
\end{figure*}

\section{Training loss implementation details}

In addition to the segmentation and recognition losses, ASSIGN has
an anticipation loss. This anticipation loss is identical to the recognition
loss, but predicts the label of the next segment. All ASSIGN variations
were trained with this anticipation loss. We show in Table \ref{tab:quantitative-anticipation-loss}
the $\mathrm{F}_{1}@k$ scores on the CAD-120 dataset for ASSIGN with
and without the anticipation loss. Similarly to previous works \cite{morais2019learning,srivastava2015unsupervised},
doing anticipation helps with the recognition results.

\begin{table}
\caption{$\mathrm{F}_{1}@k$ results for ASSIGN with and without anticipation
loss on the CAD-120 dataset.\label{tab:quantitative-anticipation-loss}}
\resizebox{\columnwidth}{!}{
\begin{tabular}{ccccccc}
\toprule 
\multirow{2}{*}{Model} &  & \multicolumn{2}{c}{Sub-activity} &  & \multicolumn{2}{c}{Object Affordance}\tabularnewline
 &  & $\mathrm{F}_{1}@0.10$ & $\mathrm{F}_{1}@0.50$ &  & $\mathrm{F}_{1}@0.10$ & $\mathrm{F}_{1}@0.50$\tabularnewline
\cmidrule{1-1} \cmidrule{3-4} \cmidrule{4-4} \cmidrule{6-7} \cmidrule{7-7} 
ASSIGN w/o anticipation loss &  & 86.2 & 71.4 &  & 91.3 & 80.4\tabularnewline
\cmidrule{1-1} \cmidrule{3-4} \cmidrule{4-4} \cmidrule{6-7} \cmidrule{7-7} 
ASSIGN &  & \textbf{88.0} & \textbf{73.8} &  & \textbf{92.0} & \textbf{82.4}\tabularnewline
\bottomrule
\end{tabular}}
\end{table}

\section{BiRNN baselines\label{app:baselines}}

We designed two baseline models for our experiments: the Independent
BiRNN and the Relational BiRNN.

The Independent BiRNN is simply a BiRNN per entity (with shared parameters
for entities of the same class) followed by an MLP to recognize the
sub-activity (or affordance) of the entity. We call it independent
because there is no message passing between the entities. More specifically,
for the $e$-th entity at the $t$-th frame we compute its BiRNN state
as
\begin{align}
h_{t}^{e} & =\mathrm{BiRNN}\left(x_{t}^{e},\overrightarrow{h}_{t-1}^{e},\overleftarrow{h}_{t+1}^{e}\right),
\end{align}
where $x_{t}^{e}$ is the frame-level feature for the $e$-th entity
at the $t$-th frame, and $h_{t}^{e}=[\overrightarrow{h}_{t}^{e};\overleftarrow{h}_{t}^{e}${]}
is the concatenation of the forward and backward hidden states produced
by the BiRNN. We recognize the label associated with the $t$-th frame
as
\begin{align}
\hat{y}_{t}^{e} & =\mathrm{Softmax}\left(\alpha\left(h_{t}^{e}\right)\right),
\end{align}
where $\alpha$ is an MLP and label recognition is done frame-wise.

The Relational BiRNN is similar to the Independent BiRNN, but it includes
\emph{inter-class} messages between the entities. The messages exchanged
in the Relational BiRNN are a mean-pooling of the hidden states of
the entities of the sender class. We compute the model BiRNN state
as
\begin{align}
h_{t}^{e} & =\mathrm{BiRNN}\left(x_{t}^{e},\overrightarrow{h}_{t-1}^{e},\overleftarrow{h}_{t+1}^{e}\right).
\end{align}
The message to entity $e$ is an \emph{inter-class} message only,
and we compute it as the average of the hidden states of the entities
of class $c^{k}$
\begin{align}
m_{t}^{inter\rightarrow e} & =\frac{1}{K}\sum_{c^{k}\neq c^{e}}h_{t}^{k},
\end{align}
where $K$ is the number of entities for which $c^{k}\neq c^{e}$.
Finally, we recognize the label associated with the $t$-th frame
as 
\begin{align}
\hat{y}_{t}^{e} & =\mathrm{Softmax}\left(\alpha\left([h_{t}^{e};m_{t}^{inter\rightarrow e}]\right)\right),
\end{align}
where $\alpha$ is an MLP. Since the Bimanual Actions dataset \cite{dreher2020learning}
is not annotated with object affordances, we do not include the human
$\rightarrow$ object message in its Relational BiRNN.

\begin{table}
\caption{\emph{Joint segmentation and label recognition} task with no pre-segmentation.
Micro and macro $\mathrm{F}_{1}$ performance on the CAD-120 dataset.\label{tab:recognition-without-ground-truth-segmentation-cad120-micro-macro}}

\centering{}\resizebox{\columnwidth}{!}{
\begin{tabular}{ccccccc}
\toprule 
\multirow{2}{*}{Model} &  & \multicolumn{2}{c}{Sub-activity $\mathrm{F}_{1}$ (\%)} &  & \multicolumn{2}{c}{Object Affordance $\mathrm{F}_{1}$ (\%)}\tabularnewline
 &  & Micro & \multicolumn{1}{c}{Macro} &  & Micro & \multicolumn{1}{c}{Macro}\tabularnewline
\cmidrule{1-1} \cmidrule{3-4} \cmidrule{4-4} \cmidrule{6-7} \cmidrule{7-7} 
Independent BiRNN &  & 58.0 & 54.2 &  & 83.3 & 73.3\tabularnewline
rCRF \cite{sener2015rCRF} &  & 68.1 & 61.3 &  & 81.5 & 77.8\tabularnewline
KGS \cite{koppula2013learning} &  & 68.2 & 66.4 &  & 83.9 & 69.6\tabularnewline
Relational BiRNN &  & 70.3 & 67.7 &  & 81.6 & 66.4\tabularnewline
ATCRF \cite{koppula2016anticipating} &  & 70.3 & 70.2 &  & 85.4 & 71.9\tabularnewline
\cmidrule{1-1} \cmidrule{3-4} \cmidrule{4-4} \cmidrule{6-7} \cmidrule{7-7} 
ASSIGN &  & \textbf{74.8} & \textbf{73.3} &  & \textbf{86.9} & \textbf{79.6}\tabularnewline
\bottomrule
\end{tabular}}
\end{table}

\begin{table}
\caption{\emph{Joint segmentation and label recognition} task with no pre-segmentation.
Micro and macro $\mathrm{F}_{1}$ performance on the Bimanual Actions
dataset.\label{tab:recognition-without-ground-truth-segmentation-bimanual-micro-macro}}

\centering{}\resizebox{.65\columnwidth}{!}{
\begin{tabular}{cccc}
\toprule 
\multirow{2}{*}{Model} &  & \multicolumn{2}{c}{Sub-activity $\mathrm{F}_{1}$ (\%)}\tabularnewline
 &  & Micro & \multicolumn{1}{c}{Macro}\tabularnewline
\cmidrule{1-1} \cmidrule{3-4} \cmidrule{4-4} 
Dreher $\etal$ \cite{dreher2020learning} &  & 64.0 & 63.0\tabularnewline
Independent BiRNN &  & 76.7 & 74.8\tabularnewline
Relational BiRNN &  & 80.3 & 77.5\tabularnewline
\cmidrule{1-1} \cmidrule{3-4} \cmidrule{4-4} 
ASSIGN &  & \textbf{82.3} & \textbf{79.5}\tabularnewline
\bottomrule
\end{tabular}}
\end{table}

\section{Frame-level micro and macro $\mathrm{F}_{1}$ results\label{app:results-micro-macro}}

We report in Section~\ref{subsec:Joint-task} the $\mathrm{F}_{1}@k$
metric for the \emph{joint segmentation and label recognition} task.
To provide a complete analysis of ASSIGN and related methods, we include
here the micro and macro $\mathrm{F}_{1}$ results on both CAD-120
(Table~\ref{tab:recognition-without-ground-truth-segmentation-cad120-micro-macro})
and Bimanual Actions (Table~\ref{tab:recognition-without-ground-truth-segmentation-bimanual-micro-macro})
datasets.

For both datasets, ASSIGN attains superior performance when compared
to related methods and baselines. It is interesting to note that the
Relational BiRNN has lower macro $\mathrm{F}_{1}$ scores than ATCRF,
even though it has higher $\mathrm{F}_{1}@k$ scores. This relates
to the discussion in Section~\ref{subsec:experimental-settings}
that frame-level micro/macro scores are not the most appropriate metric
when dealing with \emph{joint segmentation and recognition} problems.
For example, in a situation where a method over-segments a long segment,
this might not reflect badly on the frame-level metrics but it will
reflect badly on the $\mathrm{F}_{1}@k$ metric since the model effectively
splits the long segment into many short segments.

\section{Segmentation and labeling extra qualitative comparisons}

We further illustrate the segmentation and labeling results of ASSIGN
by showing some more qualitative comparisons between ASSIGN and related
methods.

For the CAD-120 dataset, we show a \emph{making cereal} and a \emph{taking
food} activities in Figures \ref{fig:qualitative-cad120-making-cereal}
and \ref{fig:qualitative-cad120-taking-food}, respectively. For \emph{making
cereal}, we observe that ATCRF under-segments the entities, which
can happen to their model whenever their ensembling strategy agrees
on a wrong label. For the \emph{taking food} activity we observe a
mixed behavior: ATCRF over-segments halfway through the video, the
\emph{moving (}$\orangebox$\emph{)} sub-activity, and under-segments
later the \emph{closing (}$\graybox$\emph{)} sub-activity. For both
scenarios and entities in them, ASSIGN correctly segments and label
the segments.

For the Bimanual Actions dataset, we show a \emph{sawing} and a \emph{pouring}
activities in Figures \ref{fig:qualitative-bimanual-s1t9t7} and \ref{fig:qualitative-bimanual-s6t3t7},
respectively. For both activities, we observe that the biggest hurdle
for Dreher $\etal$ \cite{dreher2020learning} and the Relational
BiRNN is sustaining the prediction for long actions, which leads them
to over-segmentation issues. For example, the long \emph{hold (}$\pinksalmonbox$\emph{)
}and the long \emph{saw (}$\lightpinkbox$\emph{)}, in Figure \ref{fig:qualitative-bimanual-s1t9t7},
are heavily over-segmented by them. ASSIGN, on the other hand, has
no issues with that.

\end{document}